\documentclass[preprint,11pt]{elsarticle}
\usepackage[utf8]{inputenc}
\usepackage{geometry}
\usepackage{caption}
\usepackage{url}
\usepackage{amssymb,amsbsy,bm,amsmath,comment}
\usepackage[flushleft]{threeparttable}
\usepackage{subcaption}
\usepackage{booktabs}
\usepackage{xcolor}
\usepackage{longtable}
\usepackage{soul}

\usepackage{graphicx}

\biboptions{round,authoryear}

\usepackage{color}
\usepackage{ifthen}
\definecolor{darkgreen}{rgb}{0,0.5,0}
\newboolean{showcomments}
\setboolean{showcomments}{true}
\newcommand{\desmond}[1]{\ifthenelse{\boolean{showcomments}}{\textcolor{red}{(Desmond says: #1)}}{}}

\begin{document}

\begin{frontmatter}
\title{DeepLogit: A sequentially constrained explainable deep learning modeling approach for transport policy analysis}
\author[add1]{Jeremy Oon\corref{cor1}}
\author[add1]{Rakhi Manohar Mepparambath\corref{cor1} }
\author[add1,add2]{Ling Feng\corref{cor2}}
\ead{fengl@a-star.edu.sg}
\cortext[cor1]{Authors contributed equally}
\cortext[cor2]{Corresponding author}

\address[add1]{Institute of High Performance Computing (IHPC), Agency for Science, Technology and Research (A*STAR), 1 Fusionopolis Way, \#16-16 Connexis, Singapore 138632, Republic of Singapore}

\address[add2]{Department of Physics, National University of Singapore, Singapore 117551, Republic of Singapore}

\begin{abstract}
Despite the significant progress of deep learning models in multitude of applications, their adaption in planning and policy related areas remains challenging due to the black-box nature of these models. In this work, we develop a set of DeepLogit models that follow a novel sequentially constrained approach in estimating deep learning models for transport policy analysis. In the first step of the proposed approach, we estimate a convolutional neural network (CNN) model with only linear terms, which is equivalent of a linear-in-parameter multinomial logit model. We then estimate other deep learning models by constraining the parameters that need interpretability at the values obtained in the linear-in-parameter CNN model and including higher order terms or by introducing advanced deep learning architectures like Transformers. Our approach can retain the interpretability of the selected parameters, yet provides significantly improved model accuracy than the discrete choice model. We demonstrate our approach on a transit route choice example using real-world transit smart card data from Singapore. This study shows the potential for a unifying approach, where theory-based discrete choice model (DCM) and data-driven AI models can leverage each other's strengths in interpretability and predictive power. With the availability of larger datasets and more complex constructions, such approach can lead to more accurate models using discrete choice models while maintaining its applicability in planning and policy-related areas. Our code is available on GitHub: \url{https://github.com/jeremyoon/route-choice/}
\end{abstract}


\end{frontmatter}


\section{Introduction} \label{Sn_intro}
In travel behavior analysis, traditional Discrete Choice Models (DCMs), rooted in random utility maximization (RUM) theory, have long been foundational. They offer a robust theoretical framework and interpretability by linking observed choices to underlying utility functions derived from the attributes of alternatives and decision-makers. However, conventional DCMs, particularly simpler forms like the Multinomial Logit (MNL) model, often rely on restrictive assumptions. These include linearity of utility functions and specific error distributions, which lead to the Independence of Irrelevant Alternatives (IIA) property. Such assumptions may not fully capture the complexity, non-linearity, and heterogeneity inherent in real-world travel choices~\citep{ben1985discrete, train2009discrete}.

The emergence of ``big data" in transportation, from smart card data to mobile phone data, has driven the adoption of Machine Learning (ML) and Deep Learning (DL) techniques. These data-driven models excel at identifying intricate patterns, handling high-dimensional data, and achieving high predictive accuracy without strong a priori assumptions about functional forms or error distributions (Wang et al., 2024). Despite their predictive power, these models are frequently criticized for their ``black-box" nature, making it challenging to infer underlying behavioral drivers or policy implications~\citep{sifringer2020enhancing}.

Data-driven machine learning models are increasingly used to model the choices of transportation system users~\citep{marra2021deep, wang2020deep, zhou2019bike, chang2019travel}. While interpretability is a significant advantage of DCMs over data-driven ML models, and a primary reason for their widespread use in transportation policy analysis for decades, DCMs are prone to specification errors. These errors arise from the difficulty in a priori specifying the underlying utility function, often leading to poor predictability and inaccurate interpretations. Conversely, ML models prioritize predictive accuracy and can leverage large volumes of growing digital datasets. However, their lack of clear interpretability limits their effective use by transport stakeholders and policymakers.

Despite their interpretability shortcomings, ML models have garnered increasing interest in transport modeling, with many studies comparing them against DCMs. Earlier comparative studies primarily focused on predictive accuracy between one or more models from each class within specific travel demand modeling applications~\citep{cantarella2005multilayer, hensher2000comparison, lee2018comparison}. Numerous studies have specifically compared the performance of DCMs and ML models in mode choice modeling~\citep{cantarella2005multilayer, hensher2000comparison, lee2018comparison, xie2003work, cantarella2005multilayer, zhang2008travel, omrani2013prediction, wang2020deep, sifringer2020enhancing, wang2021theory}. Fewer studies have extended these comparisons to car ownership modeling~\citep{paredes2017machine, mohammadian2002nested}, activity scheduling~\citep{allahviranloo2013daily, doherty2003application}, and route choice modeling~\citep{yao2020data}. However, comparisons of their performance in transit route choice applications using large volumes of automated data are less common in the literature.

Most studies comparing DCM and ML models have utilized either revealed preference~\citep{cantarella2005multilayer, lee2018comparison} or stated preference survey data~\citep{hensher2000comparison}, regardless of the application area. Similarly, initial transit route choice models adopted DCMs and were developed using travel surveys~\citep{eluru2012travel, hunt1990logit, lo2004modeling, raveau2011topological, anderson2017multimodal, kurauchi2012estimating}. With the increasing adoption of Automatic Fare Collection (AFC) systems in cities, smart card data provides valuable insights into transit route choices made by passengers. Consequently, a growing body of literature now leverages this data to model passenger transit route choice behavior~\citep{janovsikova2014estimation, van2014deduction, zhao2016estimation, nassir2019strategy, kim2020calibration}. Comparing the performance of DCM and ML approaches is particularly relevant when modeling transit route choices using high-volume, automatically collected smart card data, as ML approaches are considered to have an advantage with large datasets~\cite{wang2021comparing}.

This study aims to develop a hybrid approach that retain the behavioral insights of DCMs while benefiting from the superior predictive capabilities of DL models. The current study offers three key contributions: 1) We propose a simple two-step methodology that preserves the parameter interpretability of discrete choice models while enhancing accuracy through deep learning approaches, 2) We demonstrate the trade-off between parameter interpretability and performance improvement by defining the ``cost of parameter interpretability", and 3) Using the proposed approach, we illustrate how transit route choice modeling can leverage vast volumes of secondary data from smart cards and other land-use datasets to improve model performance while maintaining parameter interpretability.

\section{Literature review } \label{Sn_litreview}


Several attempts to combine the DCM and DL approaches can be found in the literature.
Studies like \cite{KARLAFTIS2011387}, \cite{VANCRANENBURGH2021100340} and \cite{hillel2020systematic}  provided a general discussion on the differences and similarities between DCM and ML models along with the potential of combining these approaches for travel demand modelling applications.  
Realizing the potential for a combined approach,  researchers have tried to combine the techniques from DCM and ML models and found that the model performance can be improved by this combined approach (\cite{sifringer2020enhancing}, \cite{han2022neural}, \cite{wang2021theory}).

\cite{marra2021deep} has studied deep learning models for transit route choice modelling and reported improvement in performance compared to the path size logit model. But, the authours provided no discussion on the parameter interpretability of the developed convolutional neural network model.

\cite{sifringer2020enhancing} proposed learning based discrete choice models by separating knowledge driven and data driven parts in the utility functions to keep the straight forward parameter interpretability of discrete choice models. They also found that to maintain the parameter interpretability of the knowledge driven part, the variables entering the knowledge driven and data driven parts should not be correlated. The methodology was demonstrated using synthetic and multiple survey datasets. We extend this work by applying their methodology to a secondary dataset that is automatically generated and available in large volume. Although learning based methods can improve the models developed using smaller datasets or synthetic datasets that are generated using simpler data generation models, they become more powerful in learning complex interactions when large volume of data is available. Hence we believe research in this direction is much needed.\par

TasteNet-MNL model was proposed by \cite{han2022neural} as an extension of the L-MNL model \citep{sifringer2020enhancing} by learning taste variations by including the interactions between individual characteristics and alternative attributes. This method may not be applicable when the individual characteristics are not available in the data, which is most likely the case with the secondary datasets that are automatically collected, like transit smart card data. \par

\cite{wong2021reslogit} introduced ResLogit, a novel deep learning-based model that seamlessly integrates a Deep Neural Network architecture into a multinomial logit model for data-driven choice modeling. Their key contribution lies in extending the systematic utility function to incorporate non-linear cross-effects using residual layers and skip connections, enabling the model to account for choice heterogeneity and complex interactions. The model offers competitive predictive performance against traditional neural networks while maintaining interpretability similar to a Multinomial Logit model. Similar to other studies~\citep{sifringer2020enhancing, han2022neural} that tried combining the ML models with DCM models, one major limitation of the study is the lack of an efficient method for hyperparameter tuning, thereby affecting the reliability and reproducibility of the results. \par

Here, we adopt a unified approach to integrate the interpretability of Discrete Choice Models (DCMs) into more complex, higher-order Machine Learning (ML) models. Specifically, we leverage the widely used Multinomial Logit (MNL) model and construct its mathematical equivalent using neural networks with convolutional layers, known as Convolutional Neural Networks (CNNs)—a popular architecture in computer vision.  Our investigations indicate that existing approaches, such as those proposed by \cite{sifringer2020enhancing} and  extended by \cite{han2022neural}, struggle to yield stable and interpretable parameters in applications like transit route choice modeling with smart card data. This instability may stem from challenges in hyperparameter tuning, underscoring the need for an approach that can deliver stable, interpretable parameter estimates alongside improved model performance for such applications.

\par

\section{Datasets and choice set generation} \label{Sn_data}
Transit smart card data from the Automatic Fare Card (AFC) system in Singapore is used in the study to estimate the transit route choice models. In Singapore, the fare payment for the majority of the public transit trips are made through the smart card known as the EZlink card. The public transport network in Singapore consists of heavy rail (Mass Rapid Transit or MRT), light rail (Light Rail Transit or LRT), and bus modes---all of which uses distance-based AFC system. Public transit fare is charged based on the distance travelled and the smart card data stores the information on the boarding stop, alighting stop, transfer stop along with the time of boarding, alighting, transfers, and passenger card type (student/adult/senior). The smart card data provides the route information for a multimodal public transit journey taken by a passenger. The transit smart card data from a typical week day of February 2018 is used in the current study to estimate the  models, which roughly has 6.2 million trips that translates into 4.5 million journeys. \par

\begin{table}[h]
\centering
\caption{Route attributes}
\label{Tb_attributes}
\small
\begin{tabular}{l l }
 \toprule
Variable & Description\\
\hline
IVTT & In Vehicle Travel Time in minutes \\
NoT & Number of Transfers \\
Fare & Public transport fare in S\$ \\
WT & Walk time for transfers in minutes\\
Fare card type & Type of fare card commuter uses (Student/Adult/Senior)\\
\bottomrule
\end{tabular}
\end{table}

The raw AFC data underwent preprocessing to prepare it for model estimation. First, individual transit trips were aggregated into complete journeys using Singapore's transfer rules, which allow up to five transfers within a two-hour window and a maximum of 45 minutes for each transfer between modes~\citep{PTtransfer}. In order to estimate the choice model, we need information on the set of choices of alternative public transit routes and their attributes. The smart card data only provides information on the chosen routes, whereas the details of other routes considered by the travelers are unknown. In our implementation of choice set generation we make use of the information that is publicly available on the bus and train network. For each origin-destination pair, we generated a comprehensive choice set by enumerating all feasible route combinations across six categories: single-mode bus, bus-bus transfers, single-mode rail, bus-rail, rail-bus, and bus-rail-bus combinations. To maintain computational tractability while preserving choice diversity, we retained the five fastest routes from each category.  More details on the generation of the choice set can be found in \cite{mepparambath2023novel}. In addition to the choice set, our algorithm also estimates the important attributes associated with each route in the choice set. The details of the attributes of the routes considered in the study are listed in Table \ref{Tb_attributes}. 

All route attributes underwent log transformation to capture the non-linear relationship between attribute values and utility. In-vehicle travel time (IVTT) was converted from seconds to minutes and log-transformed, with a minimum threshold of 2 minutes applied to handle edge cases. Transit fares were processed with a minimum value of S\$0.92 (the base fare in Singapore's distance-based pricing system) before log transformation. Walking time for transfers was incremented by one second to avoid log(0) issues and converted to minutes. The number of transfers was incremented by one before log transformation. The passenger fare card type (Student, Adult, or Senior), which was one-hot encoded into a 3-dimensional binary vector, was also used as a feature in our model. 

As the built environment is crucial in shaping how people move \citep{FARINLOYE201926}, we incorporated geographic context by encoding 30 zoning categories (as described in Table~\ref{geo_attributes}) of origin, destination, and transfer stations based on Master Plan 2014 made by Singapore's Urban Redevelopment Authority. It contains detailed plans on guiding the development of land and property in Singapore and is a good proxy for existing developments during the period of the AFC data.

\begin{table}[h]
\centering
\caption{Zoning Categories}
\label{geo_attributes}
\footnotesize
\begin{tabular}{l l l}
\toprule
Utility & Open Space & Place of Worship \\ Port/Airport & Business 2 & Sports \\
Recreation & Waterbody & Agriculture \\
Special Use & Commercial & Residential \\
Transport Facilities & Commercial \& Residential & Civic \& Community Institution \\
Health \& Medical Center & Residential with Commercial at 1st storey & Park \\
Mass Rapid Transit & Business 1 & Beach Area \\
Light Rapid Transit & Cemetery & Business Park\\
White & Hotel & Business 2 - White \\
Business 1 - White & Residential / Institution & Business Park - White\\
\bottomrule
\end{tabular}
\end{table}

For each station location, we created a 30-dimensional vector representing the land use composition within a 3-min walking distance, calculated as the percentage of area occupied by each of the 30 zoning categories. The geographic encoding was performed by intersecting circular buffers around stations with the land use polygons, transforming coordinates to Singapore's SVY21 projection system for accurate area calculations. Our feature engineering approach resulted in a 97-dimensional feature vector for models incorporating geographic and passenger characteristics, or a 4-dimensional vector for baseline models using only core route attributes.
\section{Methods}
\label{Sn_Methods}

\subsection {Models}
We start by reviewing the basic concepts in the random utility based discrete choice models, detailed description of these models is available in \cite{ben1985discrete} and \cite{train2009discrete}. Under utility maximization assumption, a decision maker chooses an alternative route that maximizes his or her utility. That is a decision maker $n$ chooses a route $r$ when utility for the route $U_{rn}$ is greater than the utility for all other routes in the choice set $R$. Utility is assumed to be random variable due to the inability of a modeler to completely characterize the utility function of the decision maker. Hence the utility is split into two components---a systematic part that is observable to the modeler and a random part which is not observable. $U_{rn}$ is a random variable with a systematic component, $V_{rn}$ that can be estimated using the observations on the choice behavior of the decision maker and a random error component, $\epsilon_{rn}$. $V_{rn}$ is assumed to be a function of attributes of the alternatives and the decision maker's characteristics, $V_{rn} = \beta X_{rn}$, where $\beta$ is the vector of parameters to be estimated and $X_{rn}$ is a vector that describes both attributes of the alternatives and decision maker's characteristics. 
\begin{gather}
\label{Eq_U_general}
U_{rn} = \beta X_{rn} + \epsilon_{rn} 
\end{gather}
The vector of parameters $\beta$ is estimated by minimizing the negative log-likelihood function given by:
\begin{gather}
\label{Eq_logLikelihood}
\mathcal{L} = -\sum_{n=1}^N\sum_{r\in R} y_{rn}\log[P_{rn}]
\end{gather}
where $y_{rn}$ is equal to 1 if the decision maker $n$ chooses alternative $r$ and 0 otherwise and $P_{rn}$ is the probability of decision maker $n$ choosing alternative $r$ and $N$ is the total number of decision makers in the data. \par

The model is made operational by making assumptions on the probability distribution on the error term, $\epsilon_{rn}$. Change in assumptions on this probability distribution results in different classes of choice models. Multinomial logit (MNL) model, which is the most commonly used class of discrete choice models is derived under the assumption that $\epsilon_{rn}$ is i.i.d. Gumbel distributed for all $r$. This assumption results in a closed form equation for choice probability as given below.
\begin{gather}
\label{Eq_P_MNL}
P_{rn} = \frac{e^{\beta X_{rn}}}{\sum_{q\in R} e^{\beta X_{qn}}}   
\end{gather}
More flexible specifications for the error term lead to other classes of choice models like the nested logit, probit and mixed logit. One variation of MNL that is often used in route choice applications is a path size logit (PSL) model. Path size logit model aims to correct MNL's overestimation of utilities for overlapping segments that are part of multiple routes in the choice set. This is achieved by including a path size term in the utility function as:
\begin{gather}
\label{Eq_U_PS}
U_{rn} = \beta X_{rn} + \beta_{PS} \log PS_r + \epsilon_{rn} 
\end{gather}
where pathsize is given by the following equation.
\begin{gather}
\label{Eq_PS}
PS = \sum_{(i, j) \in r} \frac{c_{(i, j)}}{c_r }   \frac{1}{\sum_{r \in R} \delta_{i,j}^r}
\end{gather}
\par 
where $\delta_{i,j}^r = 1$, if link $(i,j)$ belongs to route $r$ and $0$ otherwise.

\begin{figure}[h] 
\centering
\includegraphics[width=\textwidth]{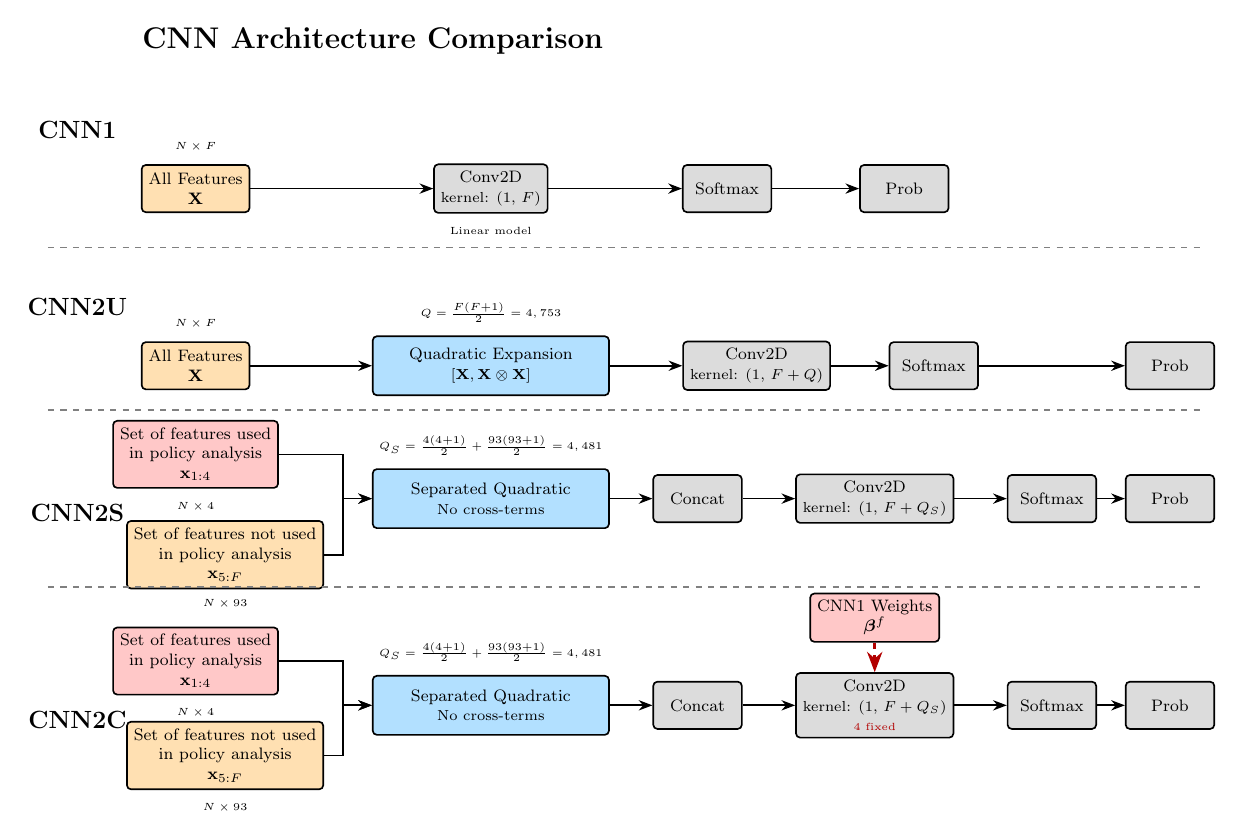}
\caption{Proposed modeling approach with CNNs.}
\label{fig:CNNarch}
\end{figure}

A linear-in-parameter MNL model can easily be implemented using a Convolutional Neural Network (CNN) with a linear filter that capture the parameters of the MNL model and one hidden layer that capture the utilities of the alternative and a softmax activation function that convert the utilities to the corresponding choice probabilities. This approach of implementing an MNL as a neural network is already applied in multiple research works~\citep{sifringer2020enhancing, han2022neural, wong2021reslogit} and is demonstrated as CNN 1 in Figure~\ref{fig:CNNarch}. In this approach, all parameters estimated in the filter ($\beta$) are equivalent to the parameters estimated in an MNL model. 

There are other existing works that are built on this CNN architecture such as the learning-based discrete choice models such as L-MNL~\citep{sifringer2020enhancing}, TasteNet~\citep{han2022neural} and ResLogit~\citep{wong2021reslogit}. In this paper we propose a simplification of the approaches in L-MNL and TasteNet models by using a fixed filter for variables for which we need interpretable parameter outputs. For other variables, which can be higher order or interaction terms of the variables we use a variable filter, values of which are estimated. The values of the fixed filter can be derived from an MNL model or the CNN equivalent of an MNL model. \par

We explored the models in an incremental fashion to reach the final model architecture, starting from the CNN 1 model, which is the MNL equivalent. In the next step, the CNN 1 model was added with quadratic terms for the features, resulting in the unconstrained quadratic CNN model (CNN 2U) as shown in Figure~\ref{fig:CNNarch}. In this model, we expanded the input features quadratically to capture pairwise interactions explicitly. This preprocessing step is crucial for the deep learning models because, unlike sequential data where positional relationships are inherent, route choice features lack natural ordering. The quadratic expansion also provides the self-attention mechanism, used later in the Transformer models, with richer representations of feature interactions. \par
Since CNN 2U model has no separation between the policy variables and other variables,  this model will result in loss of interpretability for the policy variables~\citep{sifringer2020enhancing}. Hence, in the next step we separated the policy variables from the other variables avoiding interaction terms between them, resulting in the separated quadratic CNN model (CNN 2S). We observed that CNN 2S resulted in different parameter estimates for the policy variables compared to the base CNN 1 or the MNL model as these parameters may be sensitive to the hyperparameter tuning in the DL models. Hence we decided to explore a different architecture, where the policy variables are separated and also constrained at the corresponding MNL parameter values resulting in the constrained quadratic CNN model (CNN 2C), also shown in Figure~\ref{fig:CNNarch}.

\begin{figure}[h] 
\centering
\includegraphics[width=\textwidth]{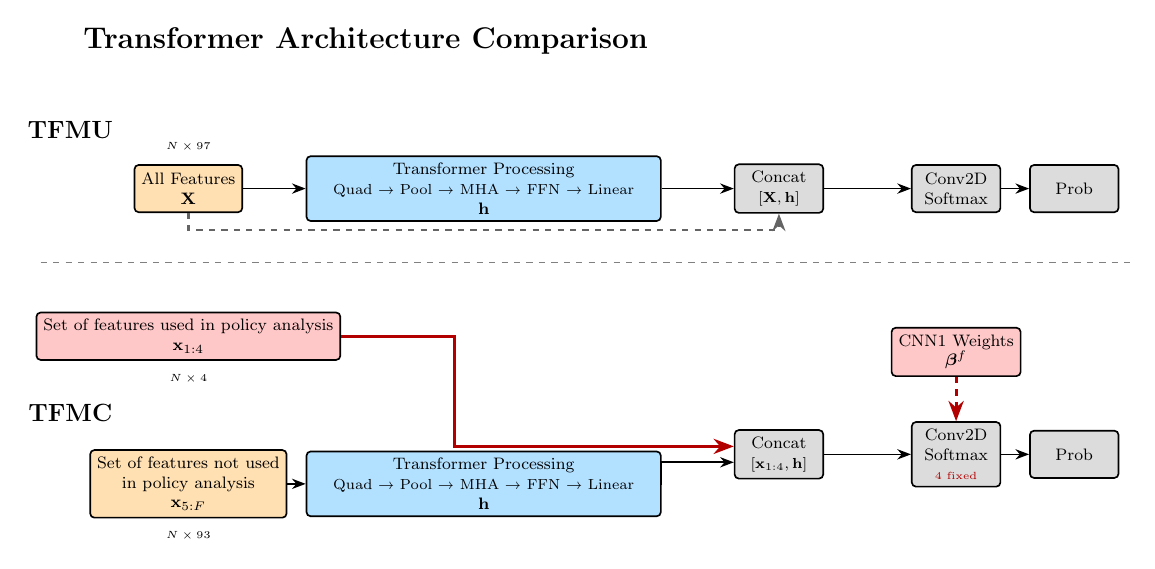}
\caption{Proposed modeling approach with Transformers.}
\label{fig:TFM arch}
\end{figure}

Following the same approach, we extend our framework to incorporate Transformer encoders, leveraging their self-attention mechanisms to capture complex feature interactions in route choice modeling, resulting in Constrained (TFM C) and Unconstrained (TFM U) versions with the Transformer architecture as shown in Figure~\ref{fig:TFM arch}.  While exploring the models with Transformers, although we explored the separated uncontrained version similar to CNN 2S, we excluded that in the list of models presented in the paper. While Transformers revolutionized sequential data processing in natural language processing and computer vision by enabling parallel computation and modeling long-range dependencies, their application to discrete choice modeling presents unique opportunities and challenges.  Before feeding features into the Transformer encoder, we apply adaptive average pooling to project the variable-length feature vectors to a fixed dimension of 512. This design choice serves multiple purposes: 1) land-use variables for areas without certain facilities often contain numerous zero-valued features. Adaptive pooling helps concentrate information from sparse features; 2) by reducing the sequence length to a manageable size, we maintain computational tractability while preserving essential information; 3) the pooling operation provides implicit normalization, which stabilizes training for the subsequent attention layers. 

Our implementation employs a 2-layer Transformer encoder with the following specifications:
\begin{itemize}
    \item \textbf{Multi-Head Attention:} 4 attention heads with dimension $d_k = d_v = 128$, allowing the model to attend to different feature relationships simultaneously
    \item \textbf{Hidden Dimension:} $d_{model} = 512$, matching the pooled feature dimension
    \item \textbf{No Positional Encoding:} Unlike sequence models, route features have no inherent position, so we omit positional encodings
    \item \textbf{Dropout:} 0.2 dropout rate for regularization
\end{itemize}
\textbf{Skip Connection:} We implement a residual connection that bypasses the Transformer layers, concatenating the original features with the transformed representations. This design ensures that: 1) basic feature information is preserved even if the attention mechanism fails to capture certain patterns; 2) the model can learn when to rely on simple features versus complex interactions; 3) gradient flow is improved during backpropagation.

Following our principle of maintaining interpretability, the constrained Transformer (TFM C) separates policy-relevant features (IVTT, Fare, Walking Time, Number of Transfers) from the Transformer processing path. These features: 1) bypass the attention mechanism entirely. 2) retain fixed weights from the pre-trained CNN 1 model; 3) are concatenated with the Transformer output only at the final layer.

\subsection{Model evaluation}
The predictive performance of both discrete choice and deep learning models is evaluated using accuracy and cross-entropy loss. Loss is a measure of the difference between the predicted value and the true observed value. Cross-entropy loss is a commonly used loss function for multi-class classification problems in neural networks and is given by:
\begin{gather}
\label{Eq_Cross-entropy}
\text{Cross entropy loss} = -\frac{1}{N}\sum_{n=1}^N \sum_{i\in C} y_{in} \log [P_{in}]
\end{gather}
which is the negative log-likelihood. Accuracy is given by the ratio of number of correct predictions to total number of predictions:
\begin{gather}
\label{Eq_accuracy}
\text{Accuracy} = \frac{\text{Number of correct predictions}}{\text{Total number of predictions}} .
\end{gather}

To avoid the potential over-fitting problem associated with neural network models, the prediction accuracy is measured using out-of-sample testing with 5-fold cross validation. For comparison, the predictive accuracies of both MNL and CNN models are evaluated using the same setup. Cross-validation is a popular method for evaluating the performance of a model on unseen data. In the 5-fold cross-validation method, the data is randomly split into five groups. One group is used as the test data and other groups together are used as the estimation or training data. The model is estimated or trained using the training set and the model's predictive performance is evaluated using the test set. The procedure is repeated five times, leaving one group out each time. The average evaluation metrics obtained across these five attempts is taken as the performance metric for the model.


To demonstrate the trade-off between the parameter interpretability and accuracy improvement, we also introduce two new terminologies, one with reference to the discrete choice models and other with reference to the deep learning models. For discrete choice models, we define the ``benefit of learning" as the improvement in the model accuracy between the pure discrete choice based approach and learning based discrete choice approach proposed in this paper. Similarly, for deep learning models, we define the ``cost of parameter interpretability" as the reduction in accuracy for deep learning models due to the constraints on the parameter values as proposed in this paper. By definition, the difference in accuracy between the pure discrete choice and deep learning models would then be equal to the sum of benefit of learning and cost of parameter interpretability.

\section{Experiment \& Results}
\label{Sn_Results}
\subsection{Training}
The discrete choice models and the dep learning  models described in Section~\ref{Sn_Methods} are estimated using the route choice observations from the transit smart card data described in Section \ref{Sn_data}. PandasBiogeme (\cite{bierlaire2018pandasbiogeme}) was used in the estimation of the discrete choice models. \par

\begin{figure}[h] 
\centering
\includegraphics[width=\textwidth]{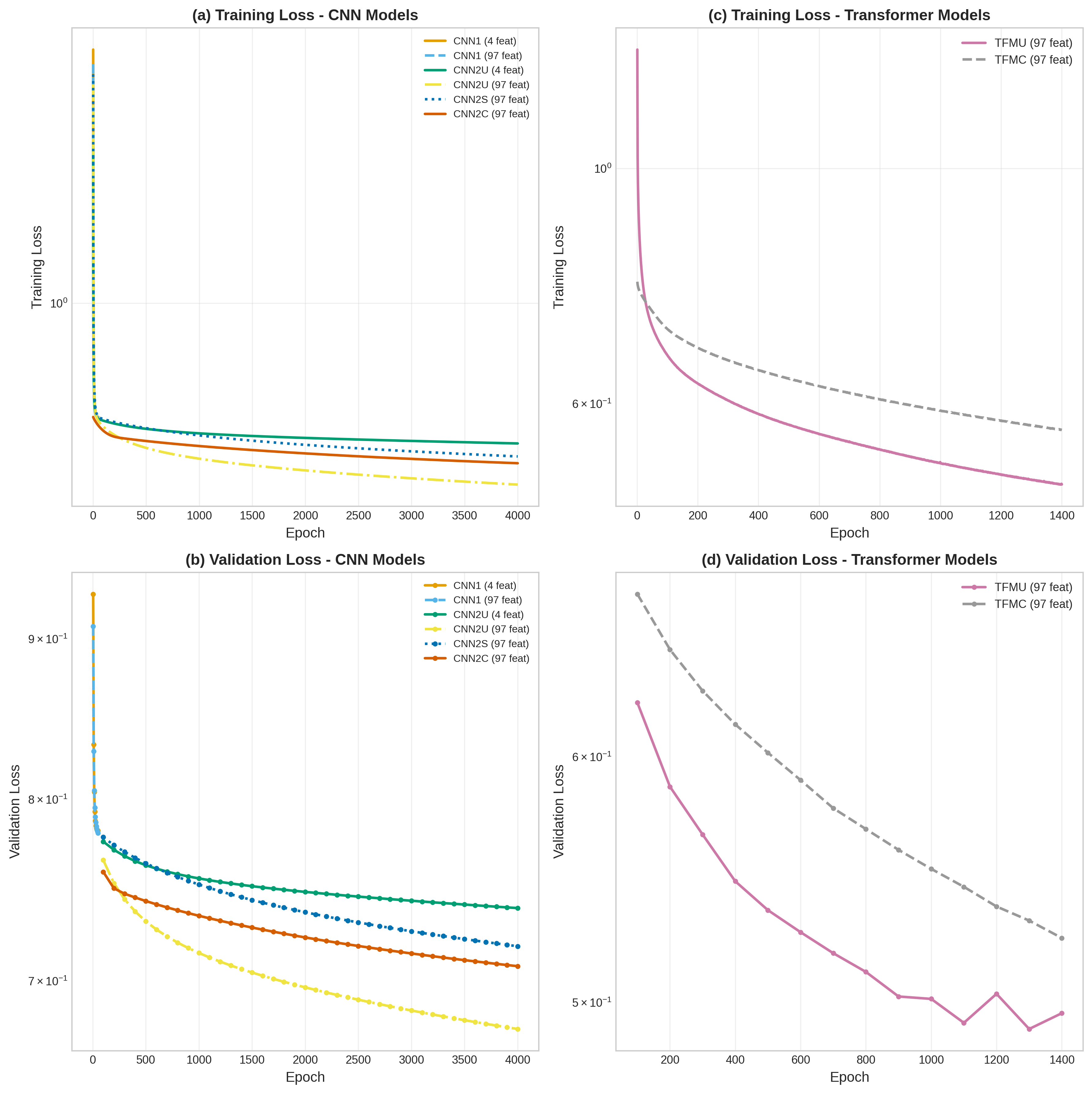}
\caption{Training and Validation loss comparison between models}
\label{fig:TrainValidloss}
\end{figure}

\begin{figure}[h] 
\centering
\includegraphics[width=\textwidth]{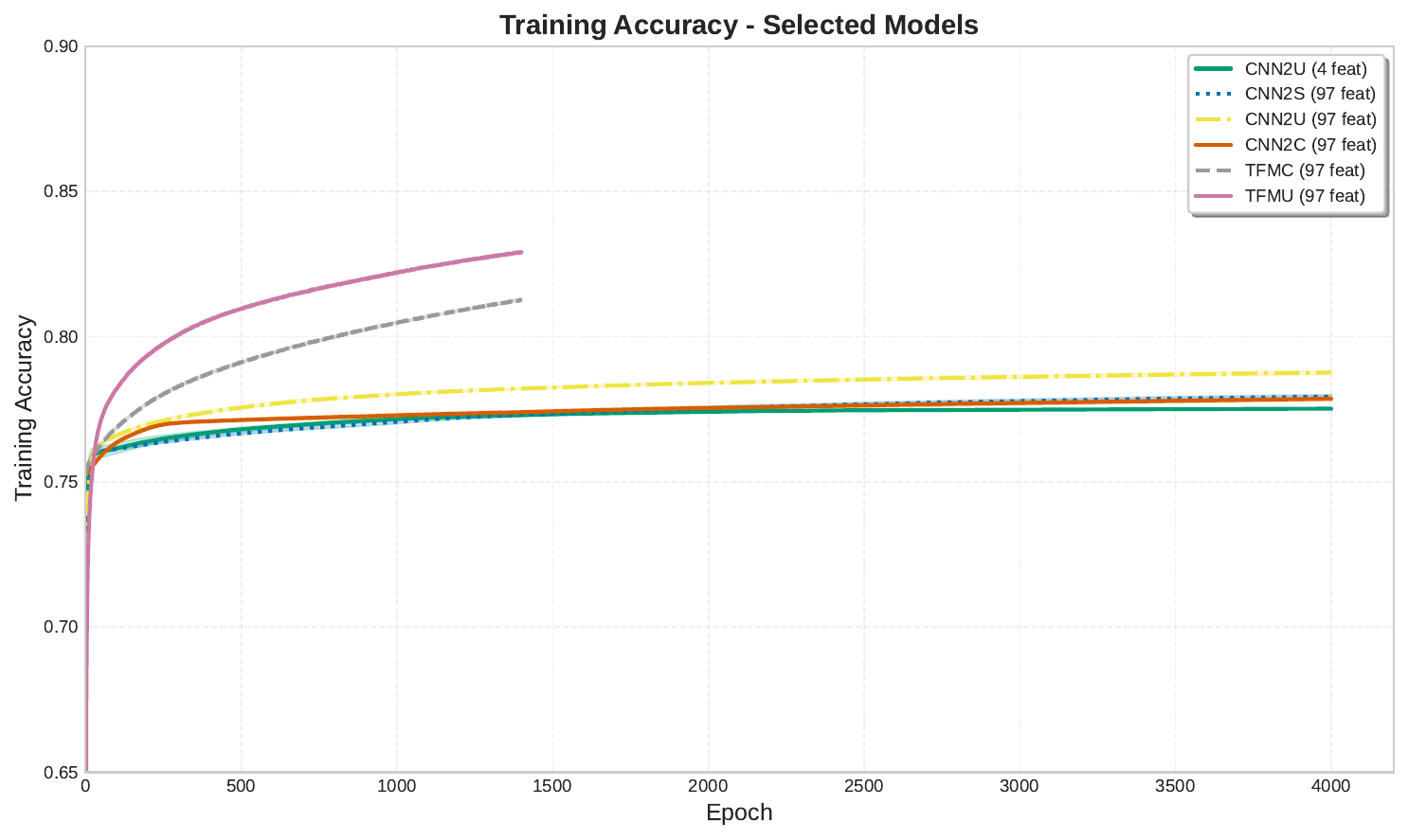}
\caption{Accuracy comparison between models}
\label{fig:accuracy_train}
\end{figure}

We used a single NVIDIA RTX6000 Ada to train all deep learning models for varying epochs: 50 for CNN 1, 4000 for CNN 2, and 1400 for Transformer (TFM) models. We observe that training for additional epochs improves both the training and validation loss and accuracy, but begins overfitting based on validation accuracy. Hence, we present results of the best validation model within the training budget. Figure~\ref{fig:TrainValidloss} presents the accuracy and loss values during training and validation, while Figure~\ref{fig:accuracy_train} presents the training accuracy for all models. As visible in the plot, Transformer models---both the constrained and unconstrained models---are able to achieve much higher accuracy within 1400 training epochs.

\subsection{Parameter estimates for discrete choice model }

 The estimation results for the discrete choice models are presented in Table \ref{Tb_MNL}. The variables used in the utility function include In-Vehicle Travel Time ($IVTT$), Walk Time for transfers ($WT$), Number of Transfers ($NoT$) and Public transit fare ($Fare$). The parameter estimates for the MNL and PSL have expected signs, with In-Vehicle Travel Time ($IVTT$), Walk Time for transfers ($WT$) and Number of Transfers ($NoT$) having negative sign. The value for the cost coefficient or the Public transit fare ($Fare$) was kept as negative one to estimate the parameter values in the willingness-to pay space. \par

As reported in multiple studies~\citep{ hensher2007development, raveau2014behavioural, kim2020calibration, mepparambath2023novel}, Number of Transfers ($NoT$) were found to have the highest negative influence on the transit route choice followed by Walk Time for transfers ($WT$) and In-Vehicle Travel Time ($IVTT$). As presented in the table, PSL model  only showed minor improvement in McFadden's ${\bar\rho}^2$ compared to the MNL model.  The pathsize variable is statistically significant as indicated by the t-statistic of the pathsize variable, but not much difference is observed between the parameter estimates and model performance of the MNL and PSL models.

\begin{table}
\centering
\caption{Parameter estimates from the discrete choice models}
\label{Tb_MNL}
\small
\begin{tabular}{l  c c c c}
 \toprule
Model   &  Parameters & Estimates  &  Std errors  & t-stat \\
\hline
Multinomial logit (MNL)   & $\beta_{IVTT}$& --2.483   &   0.020  &--120.619  \\
No. of estimated parameters = 3   &$\beta_{WT}$& --2.995  &    0.022 & --133.744\\
${\bar\rho}^2 = 0.618   $  & $\beta_{NoT}$&  --3.686 &     0.023 &  --158.906 \\
 & $\beta_{Fare}$ & --1.000	&	-	&	-\\
 \hline
 Pathsize logit (PSL) & $\beta_{IVTT}$& --2.537	&	0.021	&		--120.597 \\
No. of estimated  parameters = 4   &$\beta_{WT}$& 	--2.989 &	0.022	& --134.361	\\
${\bar\rho}^2 = 0.620  $  & $\beta_{NoT}$& 	--3.685 &	0.023	& --158.730	\\
 & $\beta_{Fare}$ & --1.000	&	-	&	-\\
  & $\beta_{Pathsize}$ & 	0.267	& 0.013& 19.326	\\
\bottomrule
\end{tabular}
\end{table}

\subsection{Parameter estimates for deep learning models}

\begin{table}
\centering
\caption{Parameter estimates from the deep learning models}
\label{Tb_NN}
\small
\begin{tabular}{l  c c c c c}
 \toprule
Model & Feature Cnt &  $\beta_{IVTT}$ & $\beta_{Fare}$  &  $\beta_{WT}$  & $\beta_{NoT}$ \\
\hline
CNN 1 & 4 & --2.453$\pm$0.009 & --1.000$\pm$0.000 & --2.890$\pm$0.005 & --3.545$\pm$0.015 \\
CNN 1 & 97 & --2.394$\pm$0.004 & --1.000$\pm$0.000 & --2.868$\pm$0.006 & --2.855$\pm$0.021 \\
CNN 2U & 4 & --1.587$\pm$0.077 & 0.684$\pm$0.147 & --1.940$\pm$0.028 & --0.760$\pm$0.172 \\
CNN 2U & 97 & --0.843$\pm$0.016 & 0.284$\pm$0.006 & --1.119$\pm$0.005 & --0.581$\pm$0.008 \\
CNN 2S & 97 & --0.709$\pm$0.018 & --1.000$\pm$0.000 & --1.596$\pm$0.046 & --0.552$\pm$0.013 \\
CNN 2C & 97 & --2.453$\pm$0.009 & --1.000$\pm$0.000 & --2.890$\pm$0.005 & --3.545$\pm$0.015 \\
TFM U & 97 & --2.256$\pm$0.027 & --0.090$\pm$0.014 & --3.048$\pm$0.025 & --0.258$\pm$0.038 \\
TFM C & 97 & --2.453$\pm$0.009 & --1.000$\pm$0.000 & --2.890$\pm$0.005 & --3.545$\pm$0.015 \\

\bottomrule
\end{tabular}
\end{table}  

Estimation results from different DL models are presented in Table~\ref{Tb_NN}. CNN 1 is the deep learning equivalent of the MNL model and as shown in Table~\ref{Tb_MNL} and Table~\ref{Tb_NN}, the CNN 1 and MNL models have very similar parameter estimates. CNN 1 model with 97 features is similar to the CNN 1 model in model structure, but with added landuse features. These additional features used in the model are neither affecting the parameter estimates of the original four attributes nor improving the model's accuracy as shown in Table~\ref{Tb_NN} and Table~\ref{Tb_Compare}. \par
The contained DL models, CNN 2C and TFM C, have the same parameter estimates as CNN 1 by design. The lack of interpretability for the unconstrained DL models, CNN 2U and TFM U, are evident from the unexpected signs of parameters in case of CNN 2U and relative change in the parameter estimates in TFM U compared to the CNN 1 models. Similar changes in parameter values were observed in case of CNN 2S model as well, which led to the selection of the proposed constrained models (CNN 2C and TFM C) for policy related applications.

\subsection{Discrete choice and deep learning model performance comparisons}\label{Sn_compare}

\begin{table}
\centering
\caption{Comparison of model performance}
\label{Tb_Compare}
\small
\begin{tabular}{l | c | c | c | c | c }
 \toprule
Model & Parameters & Train loss &Valid loss &Train acc. &Valid acc.\\
\hline
MNL 1 & 4 &  0.781$\pm$0.005 & 0.781$\pm$0.003  &  0.753$\pm$0.001  & 0.751$\pm$0.001\\
PSL & 5 &  0.781$\pm$0.005 & 0.782$\pm$0.003 &  0.752$\pm$0.001  & 0.752$\pm$0.001\\
CNN 1 & 4 &  0.782$\pm$0.000 & 0.781$\pm$0.002 & 0.752$\pm$0.000 & 0.752$\pm$0.000 \\
CNN 1 & 97 &  0.780$\pm$0.000 & 0.780$\pm$0.002 & 0.753$\pm$0.000 & 0.753$\pm$0.000 \\
CNN 2U & 14 &  0.738$\pm$0.000 & 0.738$\pm$0.002 & 0.775$\pm$0.000 & 0.775$\pm$0.000 \\
CNN 2U & 4,850 &  0.676$\pm$0.001 & 0.675$\pm$0.001 & 0.787$\pm$0.001 & 0.787$\pm$0.001 \\
CNN 2S & 4,478 &  0.718$\pm$0.001 & 0.718$\pm$0.001 & 0.779$\pm$0.001 & 0.779$\pm$0.001 \\
CNN 2C & 4,478 &  0.708$\pm$0.001 & 0.707$\pm$0.001 & 0.778$\pm$0.001 & 0.778$\pm$0.001 \\
TFM U & 3,420,258 &  0.503$\pm$0.001 & 0.496$\pm$0.014 & 0.8290$\pm$0.000 & 0.836$\pm$0.003 \\
TFM C & 3,420,165 &  0.566$\pm$0.001 & 0.524$\pm$0.002 & 0.8127$\pm$0.001 & 0.829$\pm$0.001 \\
\bottomrule
\end{tabular}
\end{table}

The performance of the discrete choice and deep learning models are presented in Table~\ref{Tb_Compare}. The results show similar train and test values for both loss and accuracy metrics for each model. The MNL, PSL, CNN 1 with 4 parameters and CNN 1 with 97 parameters have similar validation accuracy of 75\%. There is an improvement in model accuracy from 75\% to 77.5\% for the 4-feature model and 75\% to 78.7\% for the 97-feature model, when quadratic terms were introduced in the corresponding CNN 2U models. This shows the potential of using additional features like landuse in route choice model to improve the model performance and also shows that this improvement is only achieved in more complex models like the CNN 2 here and not in the regular linear-in-parameter discrete choice models similar to CNN 1. The separated and constrained CNN 2 models CNN 2S and CNN 2C have accuracy of 77.9\% and 77.8\% respectively, showing that there is not much reduction in the CNN 2 model performance when the policy parameters were constrained. Use of Transformers further improved the model performance to 83.6\% for the unconstrained TFM U model and 82.9\% for the constrained TFM C model. For Transformer models, not much difference in accuracy is observed between the constrained and constrained versions of the models.

\begin{figure}[h] 
\centering
\includegraphics[width=0.9\textwidth]{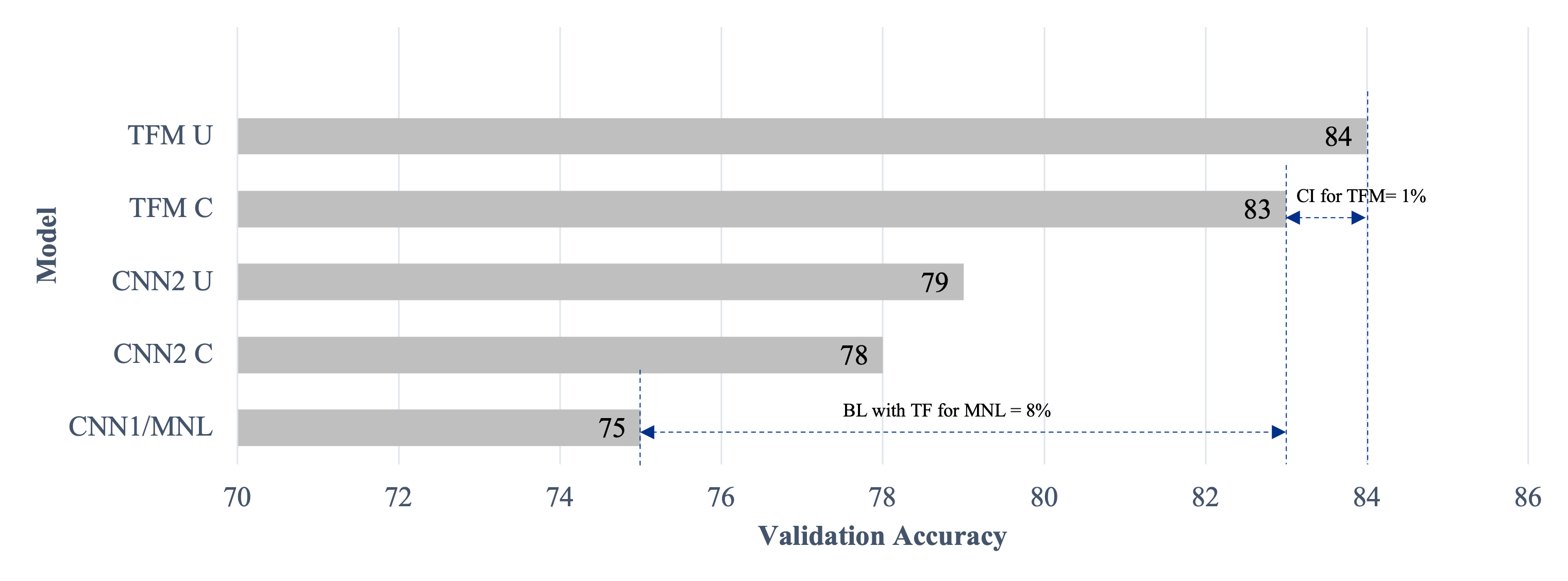}
\caption{Comparison of validation accuracy of the models. Benefit of learning (BL) with Transformer (TF) for MNL and cost of parameter interpretabilty (CI) for Transformer model (TFM) are highlighted. }
\label{fig:BL_CI}
\end{figure}

Figure~\ref{fig:BL_CI} compares the validation accuracies of various models, highlighting the ``Benefit of Learning" (BL) and ``Cost of Parameter Interpretability" (CI). The results demonstrate that the constrained Transformer model achieves the highest validation accuracy while simultaneously retaining interpretability. This constrained Transformer model improves upon the traditional Multinomial Logit (MNL) model's accuracy from 75\% to 83\%, yielding a significant 8\% ``Benefit of Learning." \par

Conversely, while an unconstrained Transformer model can achieve roughly 1\% increase in accuracy, this gain comes at a ``Cost of Interpretability", which is this additional 1\% increase in accuracy. This explicitly quantifies the trade-off between predictive performance and model transparency. Overall, these findings underscore the significant potential of hybrid approaches that integrate the robust theoretical foundation of Discrete Choice Models (DCMs) with the powerful learning capabilities of machine learning. Such integration enables the development of utility models that benefit from data-driven insights while maintaining essential interpretability for policy analysis and behavioral understanding.

\begin{table}
\centering
\caption{Summary of benefit of learning and cost of parameter interpretability}
\label{Tb_BL_CI}
\small
\begin{tabular}{l  l l l}
 \toprule
MNL Model   & &  DL model &  \\
\hline
BL with CNN& 4\%& CI for CNN& 1\%\\
BL with TF & 8\%& CI for TFM& 1\%\\
\bottomrule
\end{tabular}
\end{table}

\subsection{Elasticity Analysis} \label{Sn_Elasticity}

We analyze the results further by comparing the disaggregate point elasticities that measure the impact of an infinitesimal change of each attribute on the choice probability of the respective alternative. The disaggregate point elasticity is given by:
\begin{equation}
\label{Eq_Elasticity}
E_{x_{ni}} (i) = \frac{\partial P_n(i)}{\partial x_{ni}} \frac{x_{ni}}{P_n(i)}
\end{equation}
where $x_{ni}$ is the value of attribute $x$ for alternative $i$ and individual $n$, and $P_n(i)$ is the probability of individual $n$ choosing alternative $i$.

\begin{figure}[h] 
\centering
\includegraphics[width=\textwidth]{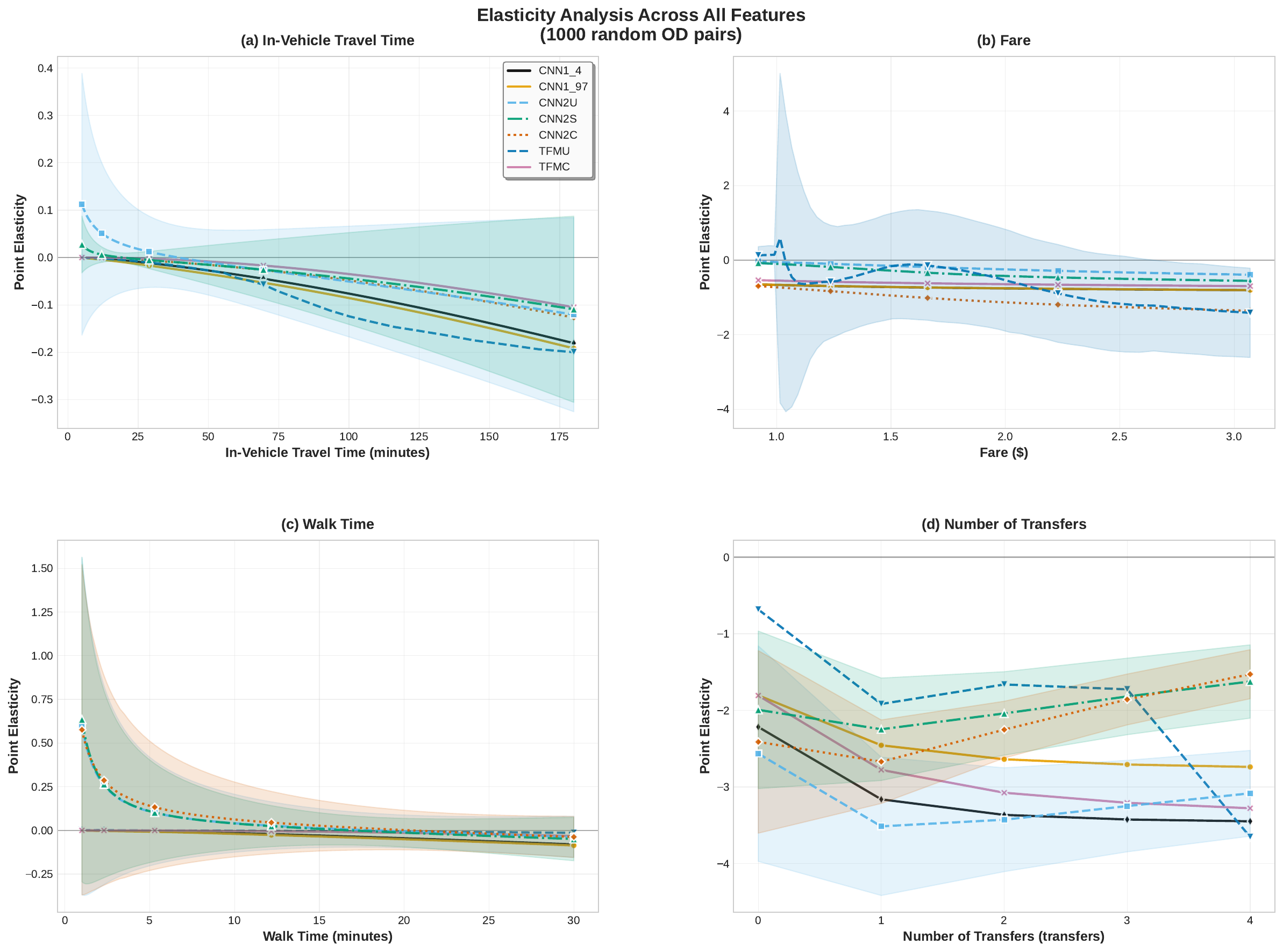}
\caption{Disaggregate point elasticity from the models}
\label{fig:Elasticities}
\end{figure}

To investigate the point elasticities systematically, we randomly selected 1,000 OD pairs where the chosen route was not the fastest option, ensuring our analysis captures behavior in competitive choice contexts. For each OD pair, we calculated point elasticities by varying the four key attributes (IVTT, Fare, Walk Time, and Number of Transfers) across their observed ranges while holding other attributes constant. The results, shown in Figure~\ref{fig:Elasticities}, reveal both expected behavioral patterns and concerning violations of theoretical principles. We provide error curves for the models with unintuitive results.

The analysis of in-vehicle travel time (IVTT) elasticities reveals unexpected patterns that warrant careful interpretation. CNN 2U and CNN 2S models exhibit positive elasticities for trips under 40 minutes, which contradicts standard behavioral theory that assumes travelers minimize journey time. This anomaly likely stems from the models capturing a specific behavioral pattern in the data: travelers who strategically extend their journeys to utilize the 45-minute free transfer window. These passengers may deliberately choose indirect routes that allow them to complete errands before reaching their final destination, rather than taking the shortest available path. The large variance observed in both models for short IVTT values supports this interpretation. In contrast, the other models maintain theoretically consistent negative elasticities, though their varying magnitudes reflect different underlying assumptions about how travelers perceive the disutility of increasing travel time.

Fare elasticity analysis uncovers a parallel anomaly in the TFM U model, which shows positive elasticities slightly above the minimum fare threshold of S\$0.92. This suggests the model predicts travelers would select more expensive options when base fares are low, a finding that defies conventional economic logic. The underlying mechanism appears similar to that observed in the IVTT analysis, where apparent preference for higher costs actually reflects strategic behavior within the fare structure.

Walk time elasticities present widespread deviations from expected behavior across multiple model architectures. CNN 2U, CNN 2S, CNN 2C, and TFM U all produce positive elasticities for walk times below 15 minutes, indicating these models predict increased utility from additional walking up to this threshold. This pattern likely captures modal preferences in the data, where travelers accept longer walks to access subway stations rather than nearby bus stops, reflecting the perceived superiority of rail transit over bus services.

Transfer elasticity patterns in CNN 2U, CNN 2C, and CNN 2S models demonstrate decreasing sensitivity to additional transfers as the number of transfers increases. Rather than maintaining consistent aversion to transfers, these models show elasticities becoming progressively less negative for multi-transfer journeys. This pattern suggests the models have identified a subset of travelers whose trip purposes differ from typical commuters---potentially leisure travelers who view transfers as less burdensome within their available time budget.

\section{Conclusions}\label{Sn_conclusion}

This  study demonstrated the significant advantages of organically integrating advanced AI architectures, such the Transformer models, into traditional econometric or choice modeling frameworks. Our findings indicate that such integrated models not only achieve superior predictive accuracy compared to conventional approaches but also crucially preserve the interpretability of key parameters. This represents a substantial enhancement, where the predictive power of sophisticated learning algorithms can be harnessed without sacrificing the behavioral or economic insights provided by transparent model parameters. In the transit route choice problem presented in this study Multinomial Logit (MNL) model, which saw its accuracy notably enhanced from 75\% to 83\% when augmented with the constrained Transformer. This quantitative ``Benefit of Learning" of 8\% underscores the powerful synergy between these methodologies, where the predictive power of advanced learning algorithms can be harnessed without sacrificing the behavioral insights provided by interpretable parameters.

Furthermore, our analysis quantifies the inherent trade-off between predictive accuracy and model interpretability. While unconstrained, highly complex models might offer marginal gains in accuracy, these often come at a measurable cost to interpretability, highlighting the inherent tension in model design. This explicit quantification of the ``Cost of Interpretability" provides a critical framework for researchers and practitioners to make informed decisions about model selection, aligning the chosen approach with the specific objectives and interpretational requirements of the research question or application. While an unconstrained Transformer model for the transit route choice problem discussed in this study can deliver a marginal 1\% increase in accuracy, this comes at a direct ``Cost of Interpretability" of 1\%, explicitly quantifying the price of relinquishing model transparency.

In conclusion, our investigations conclusively demonstrate the compelling potential of a combined modeling paradigm. This approach allows established theoretical frameworks to leverage insights from data-driven learning models, thereby significantly enhancing predictive performance. Concurrently, complex data-driven models can gain crucial interpretability by incorporating principles from well-understood, theoretically grounded methods. This hybrid strategy offers a promising avenue for developing robust, accurate, and insightful models for complex phenomena across various scientific and applied domains, ultimately facilitating more informed decision-making and policy development.

\section*{Acknowledgments}
This research is supported by the National Research Foundation, Singapore, and the Land Transport Authority under its Urban Mobility Grand Challenge Programme (Award No UMGC-L005), and Urban Redevelopment Authority under the Cities of Tomorrow R\&D Programme (Award No CoT-H1-2025-3). The views expressed herein are those of the authors and are not necessarily those of the funding agencies.

\bibliographystyle{apalike}
\bibliography{Ref.bib}

\end{document}